\documentclass{article}

\usepackage[final, nonatbib]{neurips_2022}

\usepackage[utf8]{inputenc} 
\usepackage[T1]{fontenc}    
\usepackage{hyperref}       
\usepackage{url}            
\usepackage{booktabs}       
\usepackage{amsfonts}       
\usepackage{nicefrac}       
\usepackage{microtype}      
\usepackage{verbatim}
\usepackage{blindtext}
\usepackage{tabularx}
\usepackage{tabulary}
\usepackage{threeparttable}

\usepackage[sort, numbers]{natbib}
\usepackage{xcolor}
\definecolor{navyblue}{rgb}{0.0, 0.0, 0.5}
\hypersetup{colorlinks, citecolor=blue}

\usepackage{graphicx}
\usepackage{subcaption}
\usepackage{mwe}
\usepackage{wrapfig}
\usepackage{pifont}

\usepackage{enumitem}

\usepackage{algorithmic}
\usepackage{algorithm}
\usepackage{tikz}
\usetikzlibrary{fit,calc}

\usepackage{amssymb}
\usepackage{amsmath}
\usepackage{amsthm}
\usepackage{bbm}
\usepackage{wrapfig,lipsum,booktabs}
\usepackage{dsfont}
\usepackage{mathtools}
\usepackage{mathabx}
\newlength{\depthofsumsign}
\newlength{\depthofmathbbEsign}
\usepackage{commath}
\usepackage{relsize}

\usepackage{multirow}
\usepackage{multicol}
\usepackage{caption, makecell}
\usepackage{array}
\usepackage{arydshln}
\usepackage{booktabs}
\usepackage{colortbl}
\usepackage{color}
\usepackage{tablefootnote}
\usepackage{hhline}
\usepackage{threeparttable}

\usepackage{CJKutf8}

\usepackage{comment}

\makeatletter
\def\thanks#1{\protected@xdef\@thanks{\@thanks
        \protect\footnotetext{#1}}}
\makeatother

\usepackage{ifpdf}

\theoremstyle{plain}

\theoremstyle{definition}

\title{MEDIAR: Harmony of Data-Centric and Model-Centric for Multi-Modality Microscopy}

\author{%
  Gihun Lee*\\
  KAIST AI\\
  \texttt{opcrisis@kaist.ac.kr}
  \And
  SangMook Kim*\\
  KAIST AI\\
  \texttt{sangmook.kim@kaist.ac.kr}
  \And
  Joonkee Kim*\\
  KAIST AI\\
  \texttt{joonkeekim@kaist.ac.kr}
   \AND
   Se-Young Yun$\dagger$\\
   KAIST AI\\
     \texttt{yunseyoung@kaist.ac.kr}
    \thanks{*Equal Contribution. $\dagger$ Corresponding Author.}
}

\begin{document}
\begin{CJK}{UTF8}{mj} 
\maketitle

\begin{abstract}
Cell segmentation is a fundamental task for computational biology analysis. Identifying the cell instances is often the first step in various downstream biomedical studies. However, many cell segmentation algorithms, including the recently emerging deep learning-based methods, still show limited generality under the multi-modality environment. \textit{Weakly Supervised Cell Segmentation in Multi-modality High-Resolution Microscopy Images}\footnote{\href{https://neurips22-cellseg.grand-challenge.org/}{https://neurips22-cellseg.grand-challenge.org/}} was hosted at NeurIPS 2022 to tackle this problem. We propose MEDIAR, a holistic pipeline for cell instance segmentation under multi-modality in this challenge. MEDIAR harmonizes data-centric and model-centric approaches as the learning and inference strategies, achieving a \textbf{0.9067  F1-score} at the validation phase while satisfying the time budget. To facilitate subsequent research, we provide the source code and trained model as open-source: \href{https://github.com/Lee-Gihun/MEDIAR}{https://github.com/Lee-Gihun/MEDIAR}.
\end{abstract}

\section{Introduction}
\label{sec:introduction}

Identifying cell organisms in microscopy images is fundamental for various biomedical applications \citep{microscopy_based_high_content, Ilastik, zerocostdl4mic, fiji, cell_segmentation_50years}. By partitioning the high-content images into the interested regions, segmenting cell instances is often the first step to extracting meaningful biological signals \citep{structured_tumor_immune, image_based_cell_profiling, whole_cell_segmentation, image_based_cell_phenotyping, deep_learning_pathology_survey, medical_image_analysis_survey}. As a typical microscopy system generates thousands of images in a session \citep{avoiding_replication_crisis, image_based_cell_profiling}, an automated computational approach enables the large-scale simultaneous comprehensive analysis \citep{deep_learning_pathology_survey, medical_image_analysis_survey, detecdiv}.

With the recent advances in deep learning (DL) in a wide range of vision tasks \citep{alexnet, mask_r_cnn, unet, gan, deep_learning}, DL methods have been widely adopted in microscopy image analysis \citep{cellpose, image_based_cell_profiling, datasciencebowl_2018, omnipose, stardist, tissuenet}, showing remarkable success. However, training the deep neural networks often requires a large number of labeled data \citep{deep_learning, dl_survey, dl_mining_bio_data, medical_image_analysis_survey}, and learning on the datasets with limited diversity leads the poor generalization of the model \citep{domain_generalization1, domain_generalization2, domainnet}. Such an issue is more enlarged in microscopy imaging datasets, where manually annotating cells is highly labor-intensive and time-consuming \citep{tissuenet, livecell}.
 
\textit{Weakly Supervised Cell Segmentation in Multi-modality High-Resolution Microscopy Images} (CellSeg Challenge) was hosted at NeurIPS 2022 to tackle this problem. By learning from the 1,000 labeled and 1,500+ unlabeled microscopy images, the competition aims to conduct cell instance segmentation for various situations. Although the images consist of various microscopy types, tissue types, and staining types, the metadata for the image (e.g., modality-related annotation) is not provided.

\clearpage
\label{intro}
\begin{figure}[t!]
    \centering
    \includegraphics[width=\textwidth]{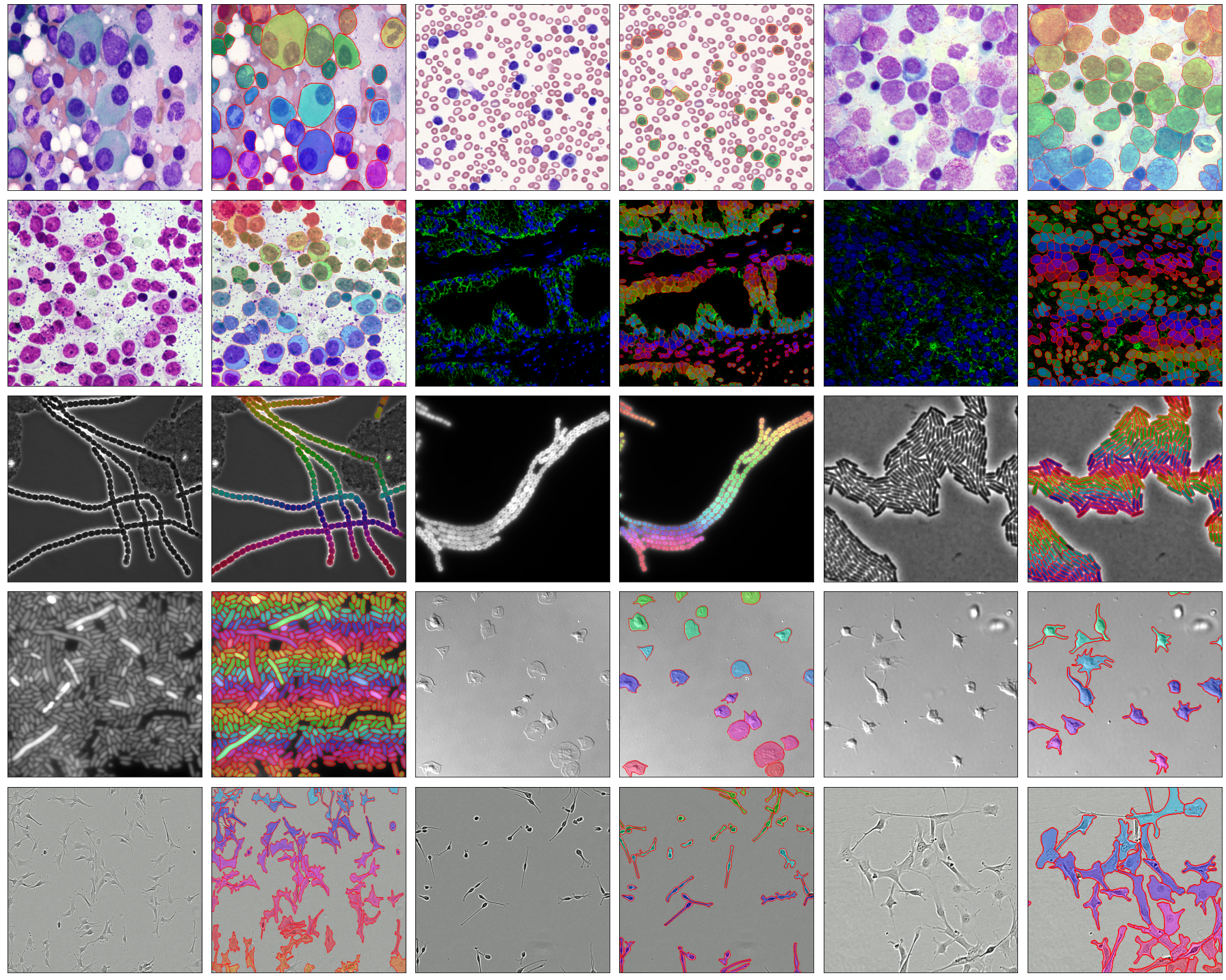}
    \vspace{-12pt}
    \caption{MEDIAR prediction results on \textit{NeurIPS 2022 CellSeg Challenge} validation images. Our proposed method identifies the cell instances evenly well across different modalities.}
    \label{fig:main_result}
\end{figure}
\vspace{-5pt}

The solution is evaluated by two criteria as follows:
\begin{itemize}
    \item \textbf{Prediction Performance: } F1-score evaluated at the IoU threshold 0.5 for the true positive.
    \item \textbf{Time Efficiency: } The running time exceeding the time tolerance:
    \vspace{-1pt}
    \begin{equation*}
    \label{eq:}
    \text{Time Tolerance}(H, W)=\begin{cases}
    			10s & \text{if} \  H \times W \leq 10^6  \\
                \frac{H \times W}{10^6} 10s & \text{if} \  H \times W > 10^6,
    		 \end{cases}
    		 \vspace{-1pt}
    \end{equation*} where $H$ and $W$ each stands for the height and width of the image.
\end{itemize}

In this paper, we propose MEDIAR, a framework to build a single generalist model for cell instance segmentation by harmonizing data-centric and model-centric approaches. On the data-centric side, MEDIAR starts from an extensive pretraining and replays data to retain the knowledge from the pretraining. Seeking balanced training towards heterogeneous modalities, MEDIAR discovers the latent modality and amplifies lacking modality samples. On the model-centric side, MEDIAR consists of a model structure with two separated heads each for cell recognition and instance distinction. To perform seamless prediction on large-scale images, we propose a stochastic test-time augmentation strategy combined with ensemble prediction. As a result, MEDIAR shows remarkable performance on a variety of microscopy images with multi-modalities while satisfying the time efficiency in the sense of tolerance budget.

To summarize, our main approaches are as follows:

\begin{itemize}
    \item We suggest an overview of the factors that make the generalization of cell segmentation difficult. Not only the different microscopy technology but also the imaging protocol, cell types, cell shapes, and even the magnification can be the source of the instance-level heterogeneity \textbf{(Section \ref{sec:hetero_modality})}
    \vspace{-1pt}
    \item We propose \textbf{MEDIAR}, a framework to conduct cell segmentation under multi-modality by a single generalist model. By combining data-centric and model-centric approaches, our method performs cell segmentation evenly well across the modalities \textbf{(Section \ref{sec:media_overview})}.
    \vspace{-1pt}
    \item In the Data-Centric view, we provide a learning strategy to balance the latent modalities and retain the knowledge from pretraining data \textbf{(Section \ref{sec:data_centric})}. 
    In the Model-Centric view, we provide a model structure to capture cell instances and a corresponding inference strategy to conduct prediction for high-resolution images under the time budget \textbf{(Section \ref{sec:model_centric})}.
    \vspace{-1pt}
    \item We analyze each component's effect in our approach, discuss the key factors of our success, and introduce the open problems. We further release the trained model, which achieved the F1-score \textbf{0.9067} on multi-modality microscopy challenge datasets (\autoref{fig:main_result} visualizes the prediction results), to facilitate the subsequent studies \textbf{(Section \ref{sec:experiments})}.
\end{itemize}

\vspace{-4pt}
\section{Difficulties in Multi-Modality Cell Segmentation}
\label{sec:hetero_modality}
\vspace{-8pt}
\begin{figure}[ht!]
    \centering
    \includegraphics[width=0.9\textwidth]{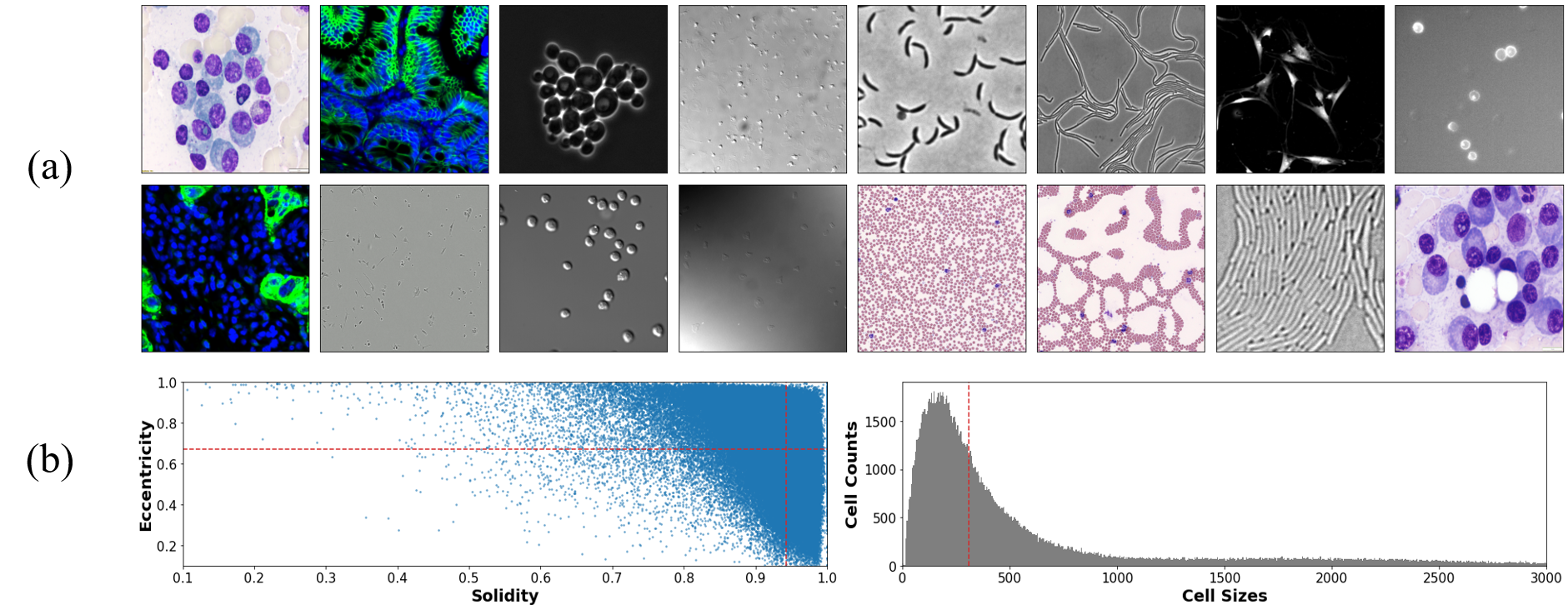}
    \vspace{-8pt}
    \caption{An overview of microscopy images in the CellSeg datasets. (a) images from various modalities. (b) statistics of the cells. The red dotted lines stand for the median value of each measure.}
    \label{fig:cell_statistics} 
\end{figure}
\vspace{-8pt}

\paragraph{Microscopy types \& Tissue types} Although the first source of multi-modality in microscopy images is the difference in the microscopy technologies used for the imaging (e.g., brightfield, fluorescent, phase-contrast, and differential interference contrast), the modality may broadly differ by the tissue type. We visualize the examples of different modalities in \autoref{fig:cell_statistics}\textcolor{red}{(a)}. Sometimes the two sources of image-level heterogeneous modality are highly correlated, as a specific microscopy technology can have an advantage in observing particular tissue types.
\vspace{-1pt}

\paragraph{Cell shapes and sizes} Another source of modality originates from cell types, which implies instance-level heterogeneity. To explore the distribution of cell shapes, we use three measures (i) Eccentricity \citep{shape_stats} $= \frac{\text{Axis}_{\text{short}}}{\text{Axis}_{\text{long}}}$ (ii) Solidity \citep{shape_stats} $ = \frac{\text{Area}}{\text{Convex Area}}$, (iii) Cell Size (pixels in each cell object).  
Note that eccentricity measures the minor and major axis ratio, and solidity measures the object's density using its convex hull. As visualized in \autoref{fig:cell_statistics}\textcolor{red}{(b)}, the cells in the images have various shapes and sizes. Note that the cells belonging to the same image may have different shapes or sizes, depending on their cell phase or the microscopy magnification.
\vspace{-1pt}

\paragraph{Annotation Inconsistency} As the criteria for the cell annotation may vary across the annotators, the heterogeneity also comes at the label-level. For example, the different standards on (i) discarding the cells in the image boundary, (ii) contour shapes of the cell, (iii) cell recognition on the object, and (iv) cell boundary could be the source of annotation inconsistency. Such noise in the data labels often degrades the performance after training \citep{noise1, noise2}.
\vspace{-1pt}

\paragraph{Others} Although not directly related to the multi-modality, the cell images are sometimes contaminated or damaged during the staining and microscopy scanning, making cell recognition more challenging. 
Another critical issue that makes cell instance segmentation difficult is touching cell objects. As the cells are often closely located with other cells without explicit object boundaries, it is often hard to assign pixels with almost similar signals to each object. Those factors also need to be considered carefully in deploying the cell instance segmentation method.
\clearpage
\section{MEDIAR: Harmony of Data-centric and Model-centric Approaches}
\label{sec:media_overview}
\vspace{-14pt}

\begin{figure}[ht!]
    \centering
    \includegraphics[width=0.99\textwidth]{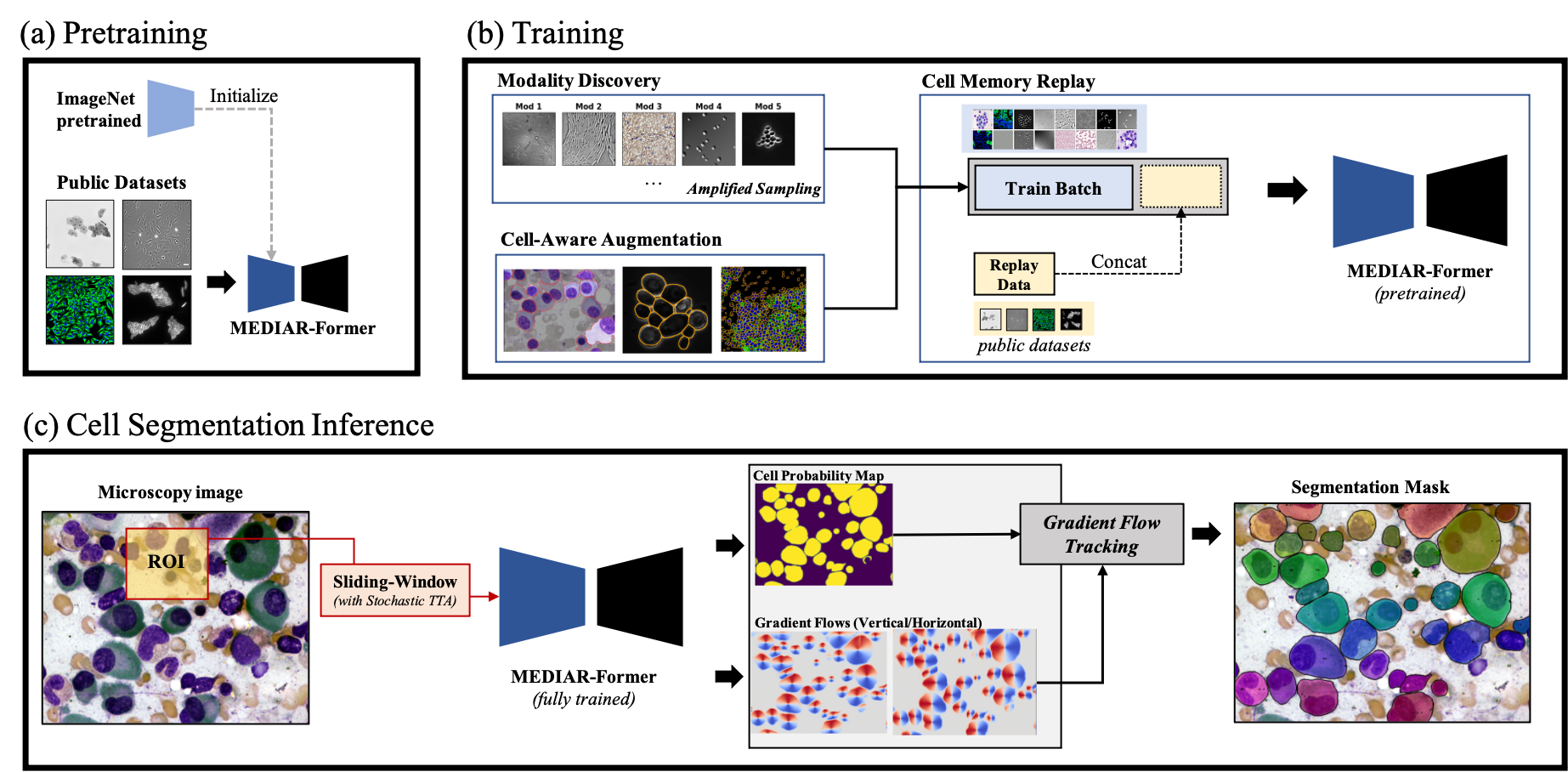}
    \vspace{-4pt}
    \caption{An overview of MEDIAR framework.}
    \label{fig:mediar_framework} 
\end{figure}
\vspace{-3pt}

The heterogeneous modalities come from various factors inevitably induce a naive approach to be felt into imbalanced learning towards specific dominant modalities. Our approach consists of two main streams to overcome this challenge: the data-centric approach and the model-centric approach. On the data-centric side, we hypothesize that (i) extensive pretraining helps capture modality-invariant features in the cell, and (ii) learning on the modality-balanced data improves generalization across modalities. On the model-centric side, we hypothesize that (i) we can train the generalist model by reducing interference between different modalities and (ii) merging multiple predictions using ensemble and test-time augmentation results in better generalization.

We illustrate an overview of the pipeline of our approach in \autoref{fig:mediar_framework}, which harmonizes the data-centric and model-centric approaches. Each phase in the pipeline ((a) pretraining, (b) training, (c) cell segmentation inference) is closely related to the remarkable performance of our proposed method MEDIAR. Note that although our framework consists of several components in each phase, they are mostly orthogonal. This implies that one can modify a part of our framework for further improvement without performance degradation due to their dependencies. In the following sections, we provide the details of key components in our proposed framework and how they contribute to the performance.
\section{MEDIAR - Data-centric Approaches}
\label{sec:data_centric}
\vspace{-6pt}
\subsection{Cell-Aware Augmentation}
\begin{figure}[ht!]
    \centering
    \includegraphics[width=\textwidth]{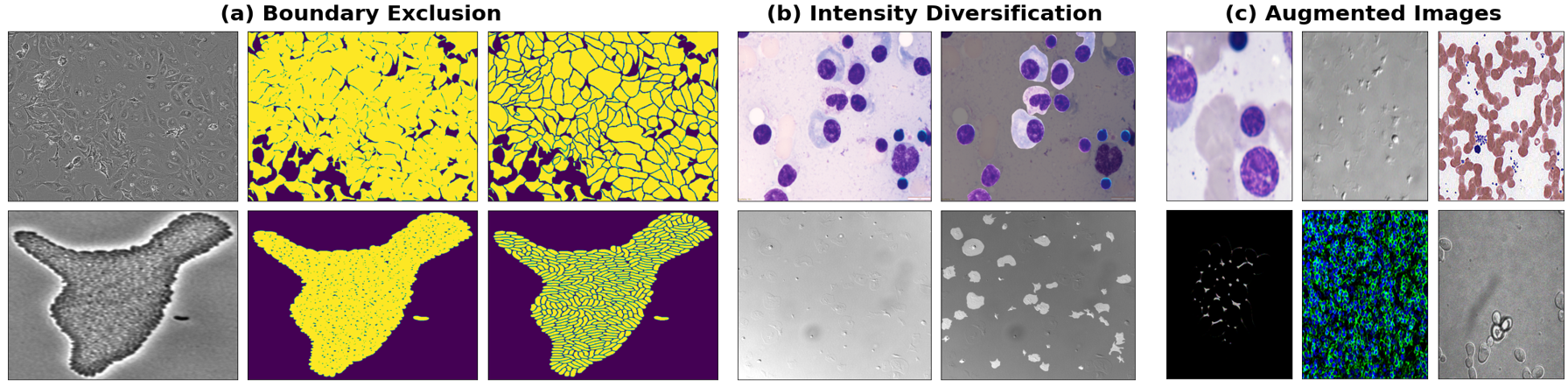}
    \caption{Examples of the proposed policies and augmented images.}
    \label{fig:cellaware_aug} 
\end{figure}

\vspace{-3pt}
MEDIAR starts from using a intensive combination of data augmentation strategies. With the prevalent augmentation methods, we propose two novel cell-aware augmentations to improve generalization. At first, as the intensity of the cells can differ in the same image in the test time, we cell-wisely randomize the intensity in the image (Cell Intensity Diversification). Second, we excluded the boundary pixels in the label. The boundary exclusion is adopted \textit{only in the pretraining phase}. We provide the combined augmentation policy in \autoref{tab:augmentation}, and visualize the examples in \autoref{fig:cellaware_aug}.

\vspace{-5pt}

\clearpage
\begingroup
\setlength{\tabcolsep}{1.25pt} 
\renewcommand{\arraystretch}{1.07}
\begin{table}[ht!]
\small
\caption{Augmentation strategies applied in the MEDIAR.}
\label{tab:augmentation}
\centering
\begin{threeparttable}
\begin{tabular}{ll} 
\toprule
\multicolumn{1}{l}{\textbf{Strategy}}    & \multicolumn{1}{l}{\textbf{Implementation Details}}        \\ 
\hline\hline
\textbf{(P)} Clip (.)       & Clip the image pixels into percentile range [0.0, 99,5]. \\
\textbf{(P)} Normalization (.)       & Normalize the images to $N(\mu=0, \sigma=1.0)$ \\
\textbf{(P)} Scale Intensity (.)     & Scale image pixels into the range [0.0, 1.0]. \\

\hline
\textbf{(S)} Zoom (0.5)                & Zooming to the scale in the range [0.25, 1.5] using nearest interpolation.  \\
\textbf{(S)} Spatial Crop (1.0)         & Cropping images as size (512, 512) at a randomly chosen center.       \\
\textbf{(S)} Axis Flip (0.5)           & Flip the array axis which corresponds to the image channels.        \\
\textbf{(S)} Rotation (0.5)             & Spatially rotate array by 90 degrees (i.e., 90\textdegree, 180\textdegree, 270\textdegree). \\ 
\hline
\textbf{(I)} Cell-Aware Intensity (0.25) & The intensity scale of each cell is scaled to the range [1.0, 1.7] \\
\textbf{(I)} Gaussian Noise (0.25)       & Add Gaussian noise $N(\mu=0, \sigma=0.1)$ to the image.  \\
\textbf{(I)} Contrast Adjustment (0.25)  & Change image intensity by factor $\gamma \in [0.0, 2.0]$  \\
\textbf{(I)} Gaussian Smoothing (0.25)   & Smoothing with Gaussian Filter with $\sigma$=1.0      \\
\textbf{(I)} Histogram Shift (0.25)      & Non-linear transformation to the intensity histogram with three control points.   \\
\textbf{(I)} Gaussian Sharpening (0.25)  & Sharpening by Gaussian filter with $\sigma$ factor 0.5 \& 1.0 with $\alpha \in [10.0, 30.0]$.\\
\hline
\textbf{(O)} Boundary Exclusion (.)  & Map the boundary pixels in the label to the background index.\\

\bottomrule
\end{tabular}
\begin{tablenotes}
\item[*] \textbf{(P)}: Pre-processing \textbf{(S)}: Spatial Augmentation \textbf{(I)}: Intensity Augmentation \textbf{(O)}: Others
\item[*] The value in the parenthesis stands for the probability of each strategy.
\end{tablenotes}
\end{threeparttable}
\end{table}
\endgroup
\vspace{-5pt}

\subsection{Two-phase Pretraining and Fine-tuning}
\vspace{-3pt}
\paragraph{Pretraining} We use 7,242 labeled images from four public datasets for pretraining: OmniPose \citep{omnipose}, CellPose \citep{cellpose}, LiveCell \citep{livecell} and DataScienceBowl-2018 \citep{datasciencebowl_2018}. MEDIAR takes two different phases for the pretraining. In phase 1, the MEDIAR-Former model with encoder parameters pretrained on ImageNet-1k is trained on the public datasets for 80 epochs. In phase 2, the pretrained model is further trained on the joint set of public datasets and train datasets for 60 epochs.
\vspace{-6pt}

\paragraph{Fine-tuning} The two pretrained models from phase 1 and phase 2 are fine-tuned with 200 and 25 epochs for each, using the train datasets. We observed that fine-tuned model from phase 1 predicts the modalities appear only in the target datasets. On the other hand, fine-tuning from phase 2 predicts the modalities included in both the public and target datasets. MEDIAR conducts the ensemble prediction using those two models. At phase 2 fine-tuning, we relabel the images, which shows the misaligned prediction between the pretrained model and the phase 1 fine-tuned model to compensate for the possible noisy labels.

\subsection{Modality Discovery \& Amplified Sampling}
\vspace{-4pt}
By observing the datasets, we find that the number of modalities is more than four. Moreover, the distribution of the number of modalities differs, resulting in an imbalanced dataset. This data imbalance may cause degradation of the model's performance for minor modalities. 

\begin{figure}[ht!]
    \centering
    \includegraphics[width=\textwidth]{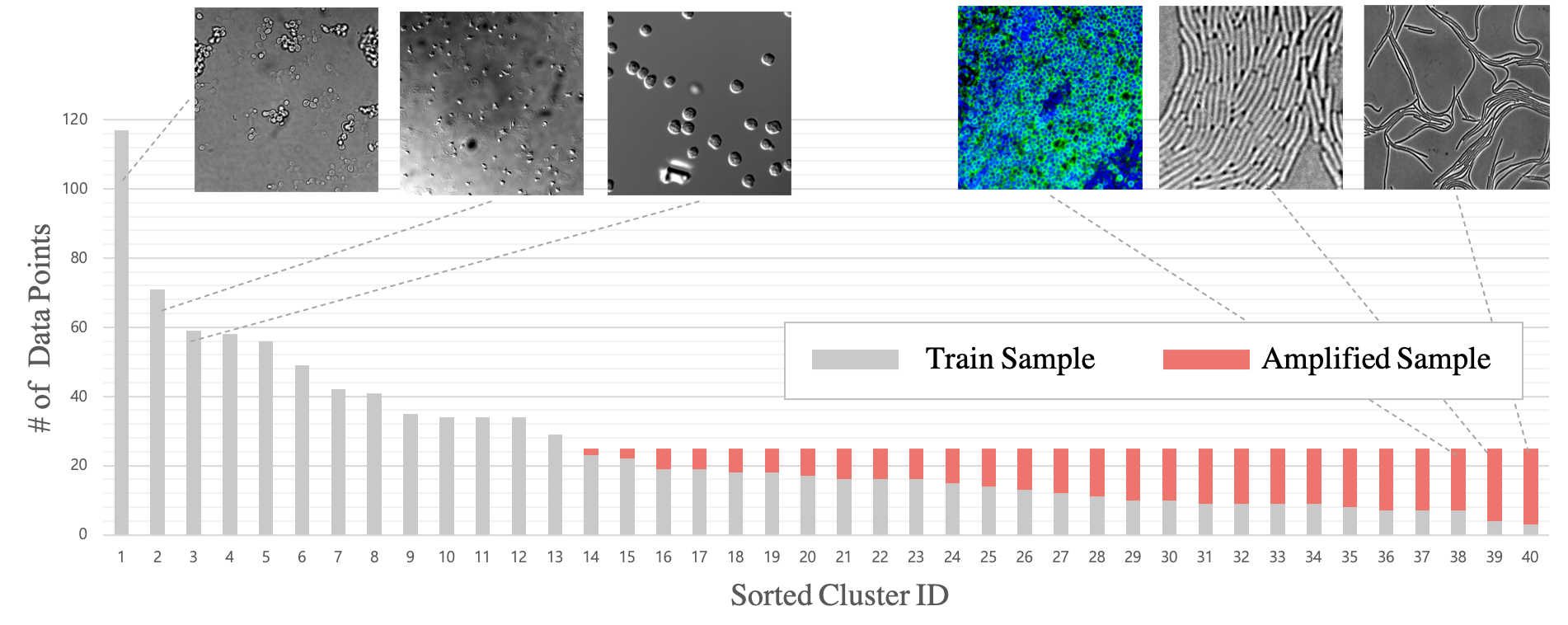}
    \caption{Discovered modalities and amplified samples in the training dataset.}
    \label{fig:modality_discovery} 
\end{figure}

We discover the latent modalities and balance their sampling ratio during training to overcome this issue. We group the encoder embeddings from the phase-1 pretrained model via the $k$-means clustering algorithm. We set the number of clusters as 40, which is large enough to sufficiently filter minor modality embeddings. To smooth the sampling ratio towards modalities, we over-sample the minor data samples. The illustration in \autoref{fig:modality_discovery} summarizes our balancing strategy.

\subsection{Cell Memory Replay}
\vspace{-3pt}
We find that the fine-tuned model performs well in most cases but degrades on some of the modalities in which the pretrained model performs well. We hypothesize that the phenomenon resembles forgetting issue \citep{cl_forgetting1, cl_forgetting2} in Continual Learning\citep{stability-plasticity}, and memory replay \citep{cl_hindsight, er_ring} mitigates the problem. We concatenate the data from the public dataset with a small portion to the batch and train with boundary-excluded labels.
\section{MEDIAR - Model-centric Approaches}
\label{sec:model_centric}
\vspace{-5pt}
\subsection{MEDIAR-Former Architecture}
\vspace{-5pt}
\begin{figure}[ht!]
    \centering
    \includegraphics[width=1.0\textwidth]{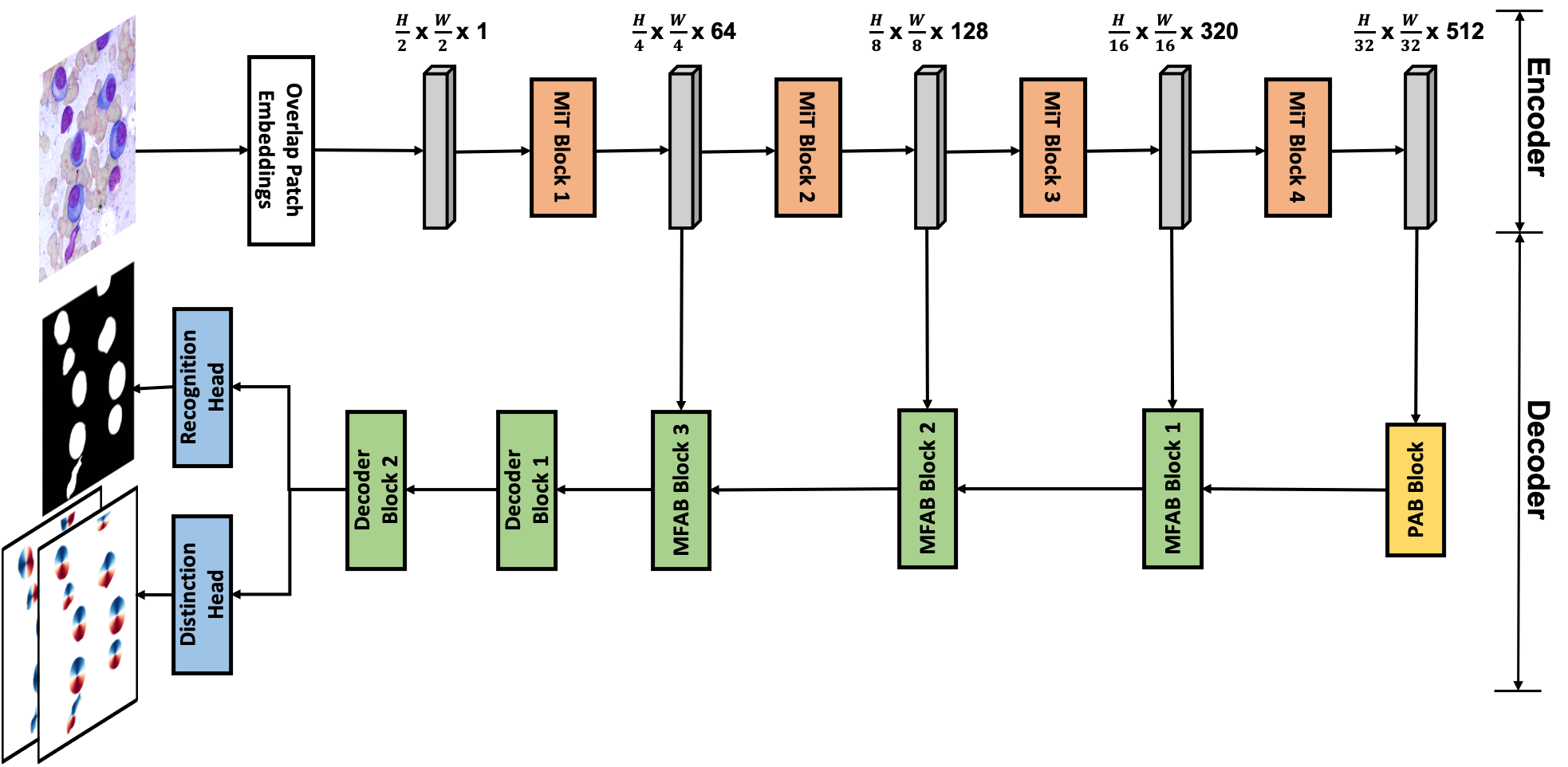}
    \vspace{-25pt}
    \caption{An overview of MEDIAR-Former model.}
    \label{fig:model_overview} 
\end{figure}

\paragraph{Encoder \& Decoder} 
MEDIAR-Former follows the design paradigm of U-Net \citep{unet}, which allows the hierarchical feature integration from the encoder to the decoder via skip connection. For the encoder and decoder, we adopt SegFormer \citep{segformer} and MA-Net \citep{manet}. The multi-scale features extracted from the encoder are concatenated through skip-connection for the decoder outputs. We use Mish \citep{mish} both in the encoder and decoder for better generalization.

The SegFormer encoder consists of multiple MiT Blocks, which include three main components as efficient self-attention, Mix-FFN, and Overlap Patch Mapping. In the efficient self-attention, the conventional self-attention \citep{attention} in ViT \citep{ViT} with the $O(N^2)$ computation complexity is reduced to $O(N^2/R)$, by modifying $K$ in the self attention $\text{Attention}(Q,K,V)=\text{Softmax}QK^T/\sqrt{d_{head}})V$, as $K = \text{Linear}(C\cdot R,C)(\hat{K})$, where $\hat{K} = \text{Reshape}(N/R,C\cdot R)(K)$. In Mix-FFN, 3x3 convolution takes place instead of positional encoding, and  Overlap Patch Mapping is recurrently applied to preserve the local continuity between patches. The Decoder MA-Net consists of two key modules, Position-wise Attention Block (PAB) for feature inter-dependencies in global view and Multi-scale Fusion Attention Block (MFAB) for semantic multi-scale feature map fusion.

\paragraph{Head} MEDIAR-Former uses two separate heads: \textit{Cell Recognition} (CR) head and \textit{Cell Distinction} (CD) head. Although the prior works use the single-head structure for each outputs \citep{cellpose, omnipose}, semantic prediction for the objects and regression for spatial gradient field interferes with each other when predicted from the same feature space, as prevalent in the Multi-Task Learning \citep{multi_task1, multi_task2}. To mitigate this issue, we separate both heads and use two 3x3 Conv-heads with BatchNorm \citep{bn}.

\clearpage
\paragraph{Learning Objective} MEDIAR-Former is learned to predict cell binary mask $y^{\text{cell}}$ from CR head ($h_{\text{CR}}$) by the binary cross-entropy loss $\mathcal{L}_{\text{BCE}}$, and cell topological maps $y^{\text{gradient}}$ from CD head ($h_{\text{CD}}$) by mean-square error loss $\mathcal{L}_{\text{MSE}}$. Here, the topological maps are normalized image gradient field generated by pseudo-diffusion as in \citep{cellpose}. The learning objective is as follows:

\begin{equation}
\label{eq:loss_term}
    \mathcal{L}(x,y) = \mathcal{L}_{\text{BCE}}(h_{\text{CR}}(f_\theta(x)),y^{\text{cell}}) + \lambda \cdot \mathcal{L}_{\text{MSE}}(h_{\text{CD}}(f_\theta(x)),y^{\text{gradient}}),
\end{equation}

where $f_\theta$ is the decoder output given the data $x$. We set $\lambda$ as 0.5 in all experiments.

\begin{figure}[t!]
    \centering
    \includegraphics[width=0.75\textwidth]{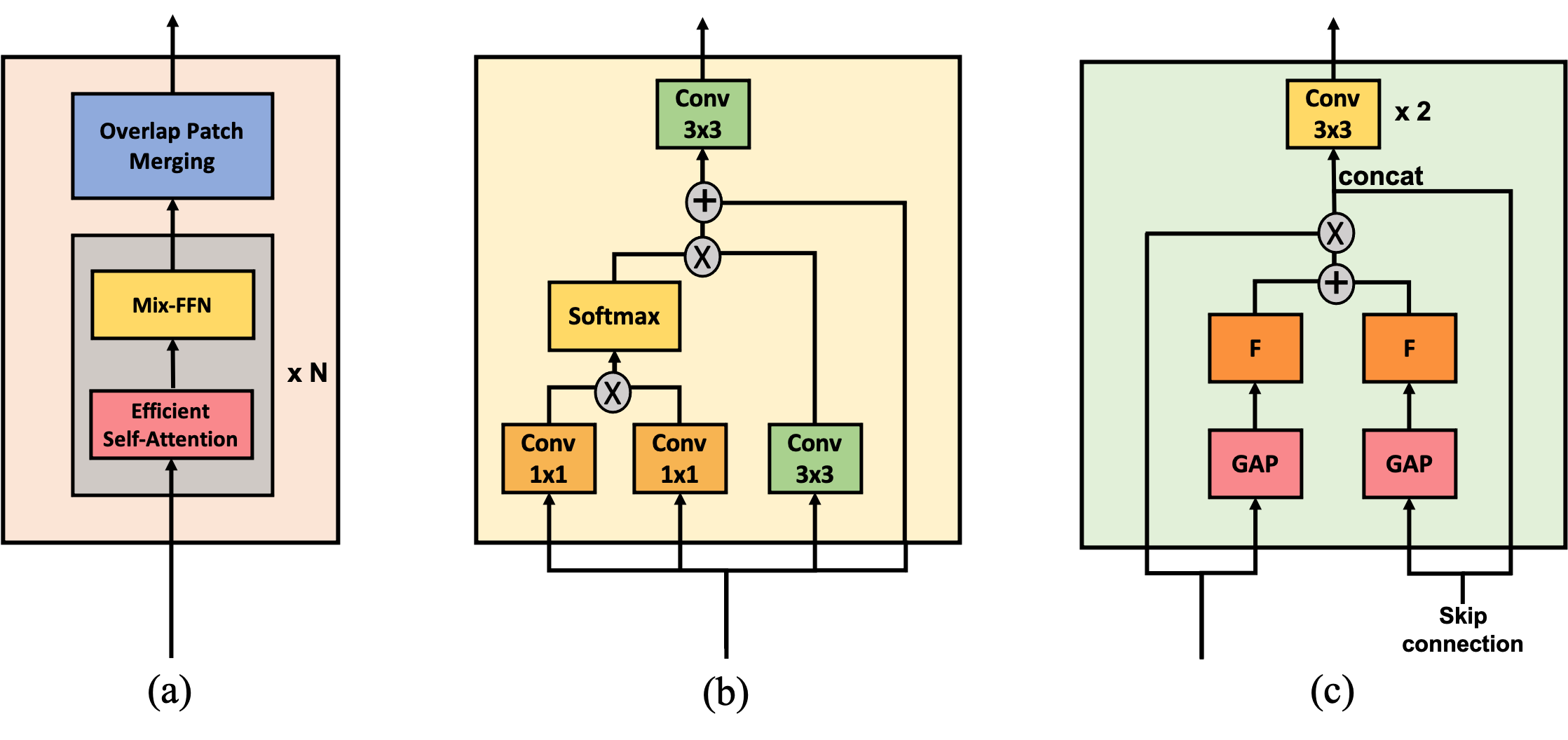}
    \caption{The details of (a) MiT Block, (b) PAB Block, (c) MFAB Block in the \autoref{fig:model_overview}.}
    \label{model_blocks}
\end{figure}

\subsection{Gradient Flow Tracking}
\vspace{-3pt}
MEDIAR adopts the gradient flow tracking in CellPose \citep{cellpose}. After filtering the cell candidates, all pixels aggregated into the cell indices by iteratively following the spatial gradient fields. First, the unit vectors are created by normalizing the gradients of each pixel. Second, the mesh grid is generated for the spatial directions, and the values are converted to the smaller starting or ending values of the unit vector directions. Third, the masks are initialized from the peak indices from the mesh grid histogram and extended until convergence. Finally, the error between pseudo-diffusion and gradient field is measured to decide whether to accept it as a cell object. We set the error threshold as 0.4. To improve time efficiency for the whole slide images, MEDIAR conducts the gradient flow tracking in a non-overlapped sliding window manner, with the patch size $2000 \times 2000$.

\begin{figure}[ht!]
    \centering
    \includegraphics[width=\textwidth]{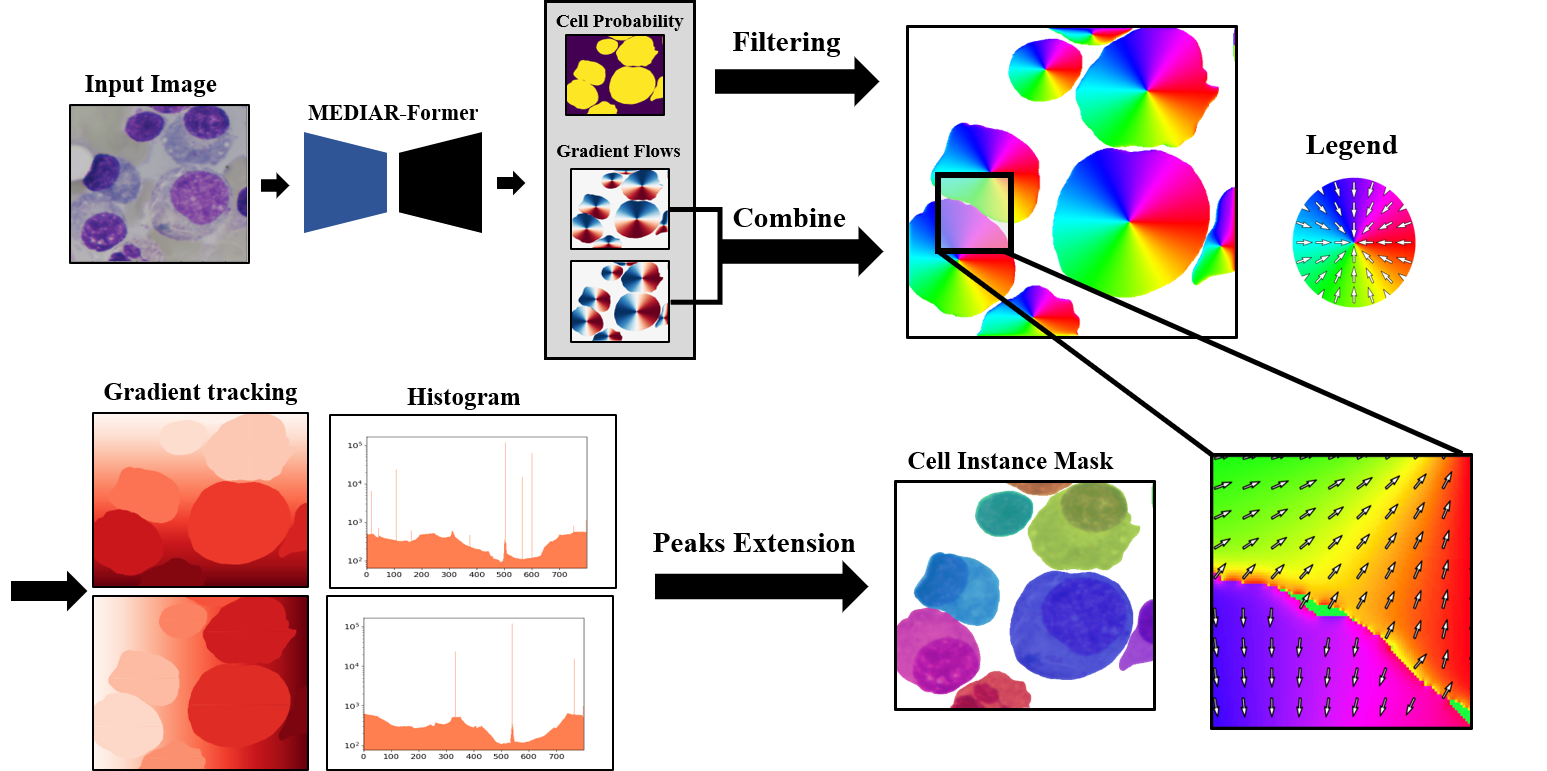}
    \caption{An example of gradient flow tracking for cell instance segmentation.}
    \label{fig:gradient tracking} 
\end{figure}

\clearpage
\subsection{Ensemble Prediction with TTA}
\vspace{-3pt}
To conduct inference on large-size input images, MEDIAR uses sliding-window inference \citep{overfeat} with the overlap size between the adjacent patches as 0.6. To predict the different views on the image, MEDIAR uses Test-Time Augmentation (TTA).
Each image is horizontally or vertically flipped, and the outputs are summed up. During each prediction, MEDIAR generates an importance map from the Gaussian Filter ($\sigma=0.125$) for each patch. Multiplying the importance map for each patch prediction output prevents recognizing the same cell at the patch boundary as multiple cells. The ensemble final prediction uses the two fine-tuned models from each pretraining phase. The modalities that the phase 1 and phase 2 models predict well are slightly different.
\section{Experiments}
\label{sec:experiments}
\vspace{-4pt}
\subsection{Experimental Setups}
\vspace{-3pt}

\paragraph{Implementation details} We use AdamW \citep{adamw} optimizer with an initial learning rate of 5e-5 in pretraining, both in the first and second phases, and 2e-5 in fine-tuning. The learning rate is decayed using cosine scheduler \citep{cosine_scheduler} using 100 interval without restarts. The code is implemented using PyTorch \citep{pytorch} and MONAI library \citep{monai} with some modifications. The base model structure and ImageNet1K-pretrained parameters are from PyTorch segmentation package \citep{smp}. We use 2 A5000 GPU cards but without the Multi-GPU training. We use mixed precision training \citep{mixed_precision} as FP-16, which reduces memory usage during training without performance degradation. We further specify the development environment in \autoref{tab:devspec} and training hyperparameters in \autoref{tabl:train_protocol}. More details are provided in the released source code.

\begingroup
\setlength{\tabcolsep}{13.0pt} 
\renewcommand{\arraystretch}{1.05}
\begin{table}[!ht]
\caption{Development environments and requirements.}
\label{tab:devspec}
\centering
\begin{tabular}{ll}
\toprule
\textbf{Environment}            & \textbf{Specification}  \\ 
\hline\hline
System                          & Ubuntu 18.04.5 LTS                            \\
\hline
CPU                             & AMD EPYC 7543 32-Core Processor CPU@2.26GHz   \\
\hline
RAM                             & 500GB; 3.125MT/s              \\
\hline
GPU (number and type)           & NVIDIA A5000 (24GB) 2ea        \\
\hline
CUDA version                    & 11.7                                   \\ 
\hline
Programming language            & Python 3.9\\ 
\hline
Deep learning framework & Pytorch \citep{pytorch} (v1.12, with torchvision v0.13.1)\\
\hline
Code dependencies       & MONAI \citep{monai} (v0.9.0), Segmentation Models \citep{smp} (v0.3.0)    \\
\hline
Specific dependencies   & ttach (v0.0.3) \citep{ttach} for Test-Time Augmentation                                         \\
\bottomrule
\end{tabular}
\end{table}
\endgroup
\begingroup
\setlength{\tabcolsep}{10.0pt} 
\renewcommand{\arraystretch}{1.02}
\begin{table*}[ht!]
\caption{MEDIAR training protocols for pretraining and fine-tuning. The epochs in the parenthesis are for the phase 2 model. Note that we include a public data sample in the fine-tuning batch.}
\label{tabl:train_protocol}
\centering
\begin{tabular}{lll} 
\toprule
\textbf{Learning Setups}            & \textbf{Pretraining} & \textbf{Fine-tuning}  \\ 
\hline\hline
Initialization (Encoder)            & Imagenet-1K \citep{imagenet}                   & from \textbf{Pretraining}                \\
Initialization (Decoder \& Head)    & He normal init                  & from \textbf{Pretraining}            \\
    Batch size                          & 9                     & 9 (with memory)                 \\
Total epochs                        & 80 (60)                   & 200 (25)                \\
Optimizer                           & AdamW \citep{adamw}                  & AdamW \citep{adamw}               \\
Initial learning rate (lr)          & 5e-5                   & 2e-5                \\
Lr decay schedule                   & Cosine \citep{cosine_scheduler} (100 interval)   & Cosine \citep{cosine_scheduler} (100 interval)  \\
Loss function                       & MSE, BCE                   & MSE, BCE                \\
\hline
Training time                       & 72 hours                   & 48 hours                \\
Number of model parameters          & 121.31 M                   & 121.31 M                 \\
Number of flops                     & 204.26 G                   & 204.26 G                \\
$\text{C{O}}_{2}eq$ \citep{carbontracker}     & 15.105g                       & 9.876g \\
\bottomrule
\end{tabular}
\end{table*}
\vspace{-6pt}
\begin{tablenotes}
\footnotesize
\item[*] * Parameters counter for pytorch models: \href{https://github.com/sovrasov/flops-counter.pytorch}{https://github.com/sovrasov/flops-counter.pytorch}
\item[*] * Flops counter for pytorch models: \href{https://github.com/facebookresearch/fvcore}{https://github.com/facebookresearch/fvcore}
\item[*] * Carbon tracker for deel learning models \citep{carbontracker}: \href{https://github.com/lfwa/carbontracker/}{Emission: https://github.com/lfwa/carbontracker/}
\end{tablenotes}
\endgroup

\paragraph{Public Datasets Usage} For the pretraining and fine-tuning, we gathered 7,242 labeled data from four public datasets as follows:
\vspace{-4pt}
\begin{itemize}
    \item \textbf{OmniPose} \citep{omnipose}: contains mixtures of 14 bacterial species. We only use 611 bacterial cell microscopy images and discard 118 worm images.
    \item \textbf{CellPose} \citep{cellpose} includes Cytoplasm, cellular microscopy, fluorescent cells images. We used 551 images by discarding 58 non-microscopy images. We convert all images as gray-scale.
    \item \textbf{LiveCell} \citep{livecell}: is a large-scale dataset with 5,239 images containing 1,686,352 individual cells annotated by trained crowdsources from 8 distinct cell types.
    \item \textbf{DataScienceBowl 2018} \citep{datasciencebowl_2018}: 841 images contain 37,333 cells from 22 cell types, 15 image resolutions, and five visually similar groups.
\end{itemize}

\begin{figure}[ht!]
    \centering
    \includegraphics[width=0.95\textwidth]{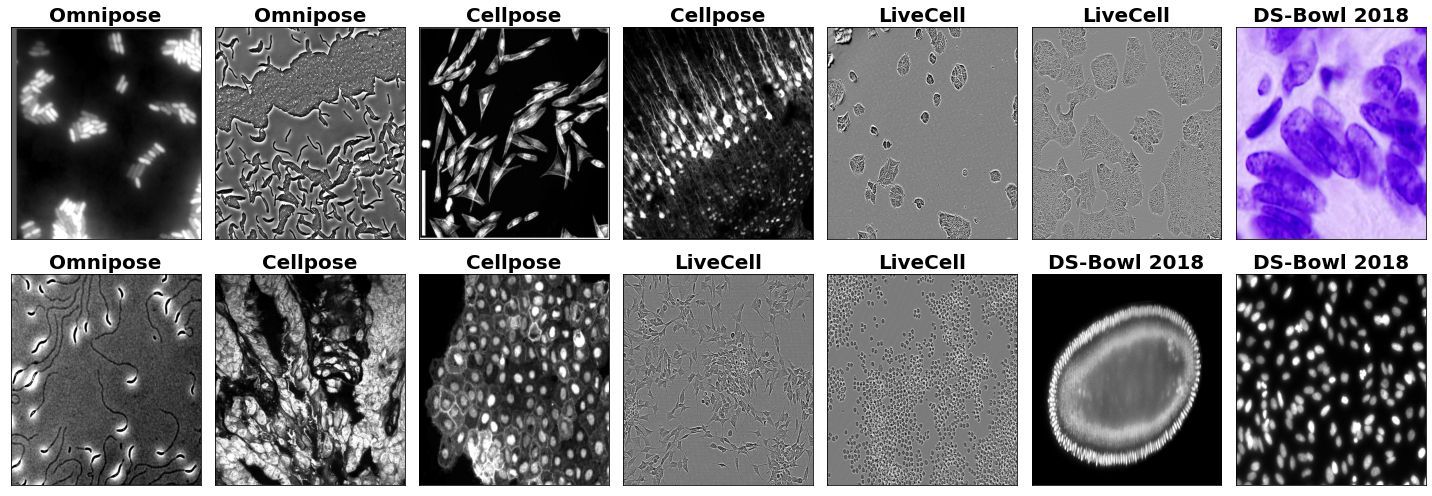}
    \vspace{-2pt}
    \caption{Example images from the collected public datasets.}
    \label{fig:public_images} 
\end{figure}

Each of the datasets contains instance-level cell mask labels. We jointly combine all the collected datasets. Example images from the collected datasets are povided in \autoref{fig:public_images}.
\vspace{-4pt}

\vspace{-10pt}
\subsection{Best Model Selection Standard}
\vspace{-4pt}
As the ground-truth label for the validation dataset is not provided, we use two different measures to select the best model checkpoint as follows:
\vspace{-3pt}
\begin{itemize}
    \item \textbf{F1-score} (\textit{with} hold-out): Randomly select 10\% of train samples as a hold-out set, and select the model that shows the best hold-out F1-score.
    \item \textbf{Cell Count} (\textit{without} hold-out): Use all data as train samples, and count the number of predicted cells on validation datasets.
\end{itemize}

\begin{figure}[ht!]
    \centering
    \includegraphics[width=\textwidth]{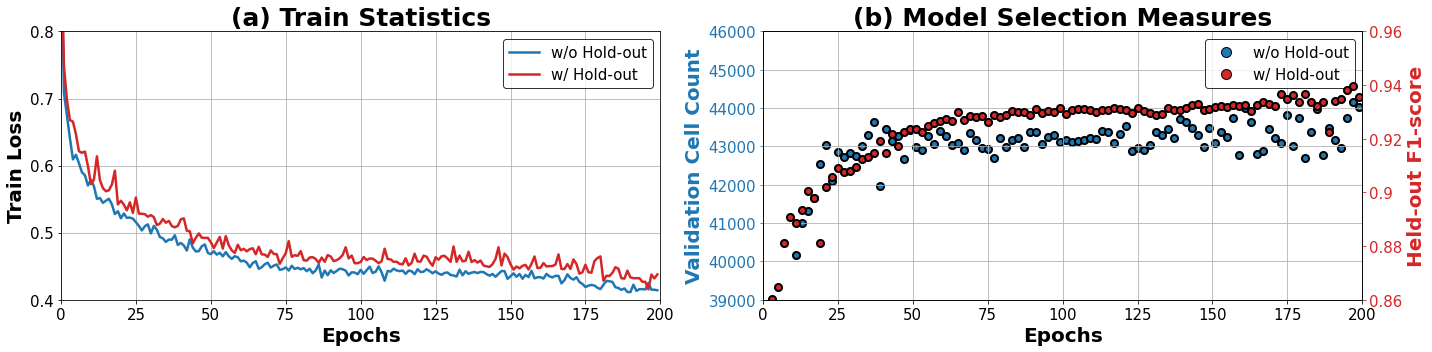}
    \vspace{-15pt}
    \caption{Change of selection measures during fine-tuning from the phase 1 pretrained model.}
    \label{fig:model_selection}
\end{figure}

The cell count measure is based on the observation that MEDIAR has its strength in sensitivity by avoiding false-positive predictions. This is because MEDIAR discards the cells above the error threshold. We plot the change of the two measures in \autoref{fig:model_selection}.
\vspace{-2pt}

\vspace{-4pt}
\subsection{Cell Instance Segmentation Performance}
\begin{figure}[ht!]
    \centering
    \includegraphics[width=0.99\textwidth]{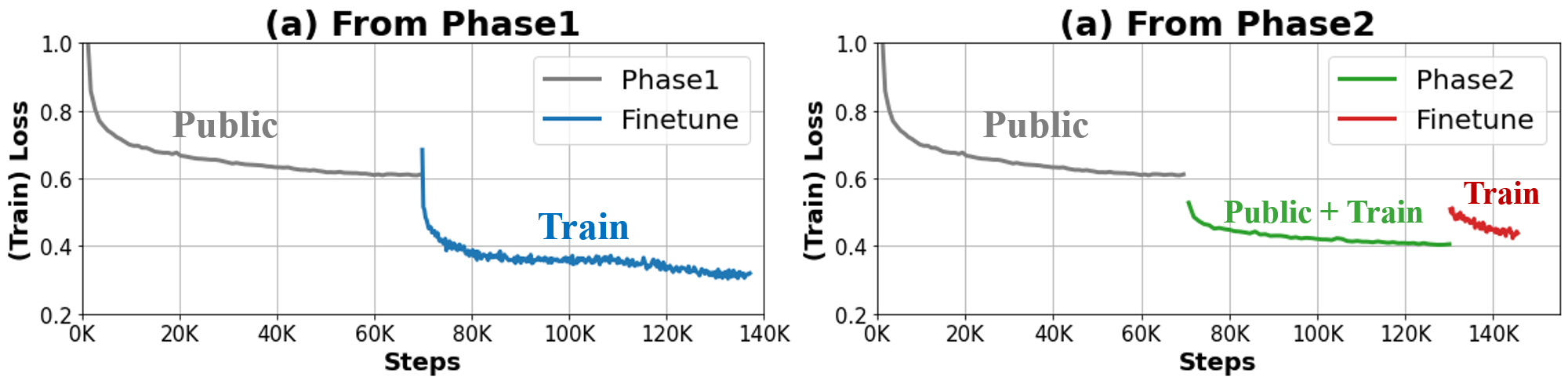}
    \vspace{-7pt}
    \caption{MEDIAR learning curves corresponding to \autoref{tab:main_performance}. Note that the learning samples for each curve are different. The text above each curve stands for the samples used in the training.}
    \label{fig:learning_curve}
\end{figure}

\vspace{-4pt}
The performance is evaluated via the F1-score at the IoU threshold of 0.5 for true positives. We use two models that each are fine-tuned from phase1 and phase2 pretraining. The validation F1-score by different inference strategies and models are in \autoref{tab:main_performance}, and learning curves for each model is plotted in \autoref{fig:learning_curve}. We use window size (512x512) with an overlap of 0.6 and a stacked error threshold of 0.4 for each cell in the inference.

\clearpage
Combined with stochastic TTA and ensemble, our MEIDAR achieves F1-score \textbf{0.9067} on validation datasets with 101 images. The prediction results on validation images are provided in \autoref{fig:main_result}. The results shows that MEDIAR predicts cells suprisingly well in various modalities. We emphasize that our MEDIAR perfectly satisfies the time limit for each image, and relaxing the time constraints would further improve the performance.
\vspace{-8pt}

\begin{table}[ht!]
\caption{MEDIAR validation F1-score by different models and inference strategies.}
\label{tab:main_performance}
\centering
\begin{tabular}{lccl} 
\toprule
\multirow{2}{*}{\textbf{Inference Strategy }} 
& \multicolumn{2}{c}{\textbf{Prediction Model }} 
& \multirow{2}{*}{\textbf{Implementation Details }}    \\
& From Phase 1 & From Phase 2   &       \\ 
\hline\hline
MEDIAR              & 0.9000            & 0.9011    & Sliding window \& Gradient tracking   \\
+ Stochastic TTA    & 0.9027            & 0.9048    & Flipping \& Importance map          \\
+ Ensemble          & \multicolumn{2}{c}{\underline{\textbf{0.9067}}}    
& Predict using the both models \\
\bottomrule
\end{tabular}
\end{table}

\vspace{-10pt}

\vspace{-5pt}
\subsection{Ablation Study}
\vspace{-3pt}

\paragraph{Model Structure} To examine the model structure's effect, we compare MEDIAR Former's performance to the model structures from prior works: U-Net \citep{unet}, Swin Unetr \citep{swin_unetr}. For a fair comparison, we train each model from scratch for 200 epochs without pretraining. The initial learning rate is set as 5e-5, and other hyperparameters are the same as the fine-tuning protocol specified in \autoref{tabl:train_protocol}. The results are in \autoref{ablation:model_structure}, which shows that MEDIAR Former significantly outperforms the others.

\vspace{-2pt}
\begingroup
\setlength{\tabcolsep}{6pt} 
\renewcommand{\arraystretch}{0.99}
\begin{table}[ht!]
\vspace{-5pt}
\caption{MEDIAR performance on different model structures.}
\label{ablation:model_structure}
\centering
\begin{tabular}{lccc} 
\toprule
\multicolumn{1}{c}{\textbf{Model Structure}} & \textbf{Model Component}                           & \textbf{(Train) Loss} & \textbf{(Valid) F1-score}  \\ 
\hline\hline 
U-Net \citep{unet}                           & \multirow{3}{*}{None}                        & 0.9592                     & 0.5473                     \\
UNetr \citep{unetr}                          &                                              & 0.6777                     & 0.6034                     \\
Swin Unetr \citep{swin_unetr}                &                                              & 0.6776                     & 0.6320                     \\ 
\hline 
\multirow{4}{*}{MEDIAR Former}               & \multicolumn{1}{l}{Base Structure}           & 0.6206                     & 0.6503                     \\
                                             & \multicolumn{1}{l}{+ Encoder Initialization} & 0.4507                     & 0.8292                     \\
                                             & \multicolumn{1}{l}{+ Decoder Scale-up}       & 0.4268                     & 0.8347                     \\
                                             & \multicolumn{1}{l}{+ Head Separation}        & 0.4144                     & \underline{\textbf{0.8424}}                     \\
\bottomrule
\end{tabular}
\end{table}
\endgroup
\vspace{-3pt}

\paragraph{Data-centric Components} In \autoref{ablation:data_centric}, we provide the effect of each data-centric approach with the details. Although the labeling consistency does not improve solely, we observe that it predicts some modalities are more robust. We combine the labeling consistency approach when fine-tuning from phase 2, which is used for the ensemble prediction. Note that all the experiments in \autoref{ablation:data_centric} use a combined augmentation strategy in \autoref{tab:augmentation}.

\vspace{-2pt}
\begingroup
\setlength{\tabcolsep}{1.5pt} 
\renewcommand{\arraystretch}{1.1}
\begin{table}[ht!]
\vspace{-5pt}
\caption{Effect of data-centric components with implementation details.}
\label{ablation:data_centric}
\centering
\begin{tabular}{lcl}
\toprule
\multicolumn{1}{c}{\textbf{Data-centric Component}}  & \textbf{(Valid) F1-score} & \multicolumn{1}{c}{\textbf{Implementation Details}}  \\ 
\hline\hline
MEDIAR (From Phase1)                    & 0.8801            & Fine-tune 200 epochs from phase 1 model \\
+ Cell-Aware Augmentation               & 0.8881            & Apply intensity diversification augmentation\\
+ Amplified Sampling                    & 0.8921           & Balanced sampling for discovered 40 modalities \\
+ Cell Memory Replay                    & \underline{\textbf{0.9000}}            & Exclude boundary for public memory data \\
+ Labeling Consistency                  & 0.8979        & Relabeling train datasets by phase2 model \\
\bottomrule
\end{tabular}
\end{table}
\endgroup

\paragraph{Cell Memory Replay} We further investigate the effect of varying memory ratios for the fine-tuning phase. As suggested in \autoref{tab:ablation_memory}, replaying only one memory data per batch is enough for preserving the knowledge from the pretraining phase, showing the sweet spot for the prediction performance.

\vspace{-8pt}
\begin{table}[ht!]
\caption{MEDIAR performance by the varying memory ratio.}
\label{tab:ablation_memory}
\centering
\begin{tabular}{cccl} 
\toprule
\textbf{Pretrained Model}          & \textbf{Train : Memory} & \textbf{(Valid) F1-score} & \multicolumn{1}{c}{\textbf{Implementation Details}}         \\ 
\hline\hline
\multirow{3}{*}{\textbf{ Phase 1}} & 9 : 0                   & 0.8801                    & \multirow{3}{*}{Fine-tuning for 200 epochs}  \\
                                   & 8 : 1                   & \underline{\textbf{0.9000}}                    &                                              \\
                                   & 7 : 2                   & 0.8881                    &                                              \\ 
\hline
\multirow{3}{*}{\textbf{Phase 2}}  & 9 : 0                   & 0.8876                   & \multirow{3}{*}{Fine-tuning for 25 epochs}   \\
                                   & 8 : 1                   & \underline{\textbf{0.9011}}                    &                                              \\
                                   & 7 : 2                   & 0.8849                    &                                              \\
\bottomrule
\end{tabular}
\end{table}

\subsection{Time Efficiency}
\begin{figure}[ht!]
    \centering
    \includegraphics[width=\textwidth]{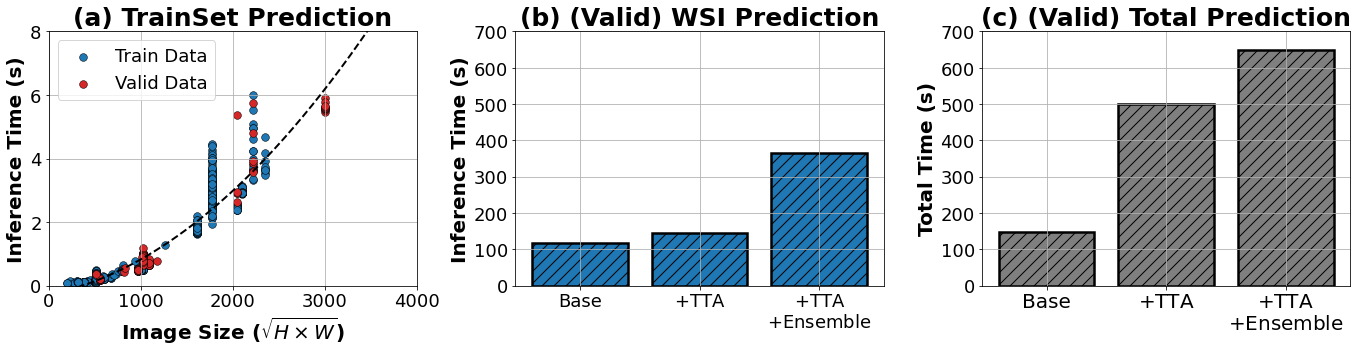}
    \caption{The time cost of the MEDIAR prediction on train datasets (1,000 images) and validation datasets (101 images). The validation datasets include one WSI.}
    \label{fig:time_efficiency}
\end{figure}

\vspace{-3pt}
We measure the time cost of MEDIAR prediction using the environment specified in \autoref{tab:devspec} with a single A5000 GPU card. The results are plotted in \autoref{fig:time_efficiency}. As suggested in \autoref{fig:time_efficiency}\textcolor{red}{(a)}, MEDIAR conducts most images in less than 1sec, and this depends on the image size. Note that although the complexity is not linear as we use the transformer structure in our encoder, its complexity $O(N^2/R)$, which is relaxed from $O(N^2)$. We also measure the total prediction time cost on 101 images in validation data, which includes one WSI with the size of $8415 \times 10496$. As in \autoref{fig:time_efficiency}\textcolor{red}{(b)} and \autoref{fig:time_efficiency}\textcolor{red}{(c)}, the total prediction time considerably increased by using the TTA strategy, but it only slightly increases the WSI prediction. Even with the TTA and ensemble, we emphasize that MEDIAR satisfies the time budget for all images.
\vspace{-5pt}

\vspace{-2pt}

\subsection{Using unlabeled images}
\vspace{-3pt}
We examine some standard semi-supervised learning approaches to exploit 1,500+ unlabeled images in train datasets. (i) \textit{Consistency Regularization}: We add a consistency loss \citep{consistency1, consistency2} term to match the prediction on the clean image and the distorted images to the model's consistency on unlabeled images. For the distortion, we use various combinations of the augmentations in \autoref{tab:augmentation}. (ii) \textit{Reconstruction Error} We add a reconstruction loss \citep{reconstruction1, reconstruction2} term by using additional head module. The model learns to reconstruct the unlabeled images. We test two reconstruction errors: one for only the pixels corresponding to the cell masks, and the other is the entire image reconstruction. (iii) \textit{Pseudo Labeling}: Using the pretrained model and fine-tuned model, we assign pseudo label \citep{pseudo_label1, pseudo_label2} to the unlabeled images to use as like labeled images. Unfortunately, all the above methods could not improve performance, thereby only the 1,000 labeled images are used in MEDIAR. Although not included in our MEDIAR, we expect that self-supervised learning approaches \citep{ssl1, ssl2, ssl3, ssl4, ssl5, ssl6} can be a promising alternative direction for using the unlabeled images.

\vspace{-6pt}
\begingroup
\setlength{\tabcolsep}{6pt} 
\renewcommand{\arraystretch}{1.1}
\begin{table}[ht!]
\caption{Approaches for using unlabeled images.}
\centering
\label{tab:unlabeled}
\begin{tabular}{c|cccc} 
\toprule
\multirow{2}{*}{\textbf{Methods}} & \multirow{2}{*}{\textbf{Baseline}} & \multicolumn{3}{c}{\textbf{Semi-Supervised Approach}}  \\
                                  &                                     & Consistency Loss & Reconstruction Loss & Pseudo Labeling            \\ 
\hline
\textbf{(Valid) F1-score}           & \textbf{\underline{0.8801}}                              & 0.8720      & 0.8798         & 0.8655                  \\
\bottomrule
\end{tabular}
\end{table}
\endgroup

\subsection{Failure cases of MEDIAR prediction}
\begin{figure}[ht!]
    \centering
    \includegraphics[width=\textwidth]{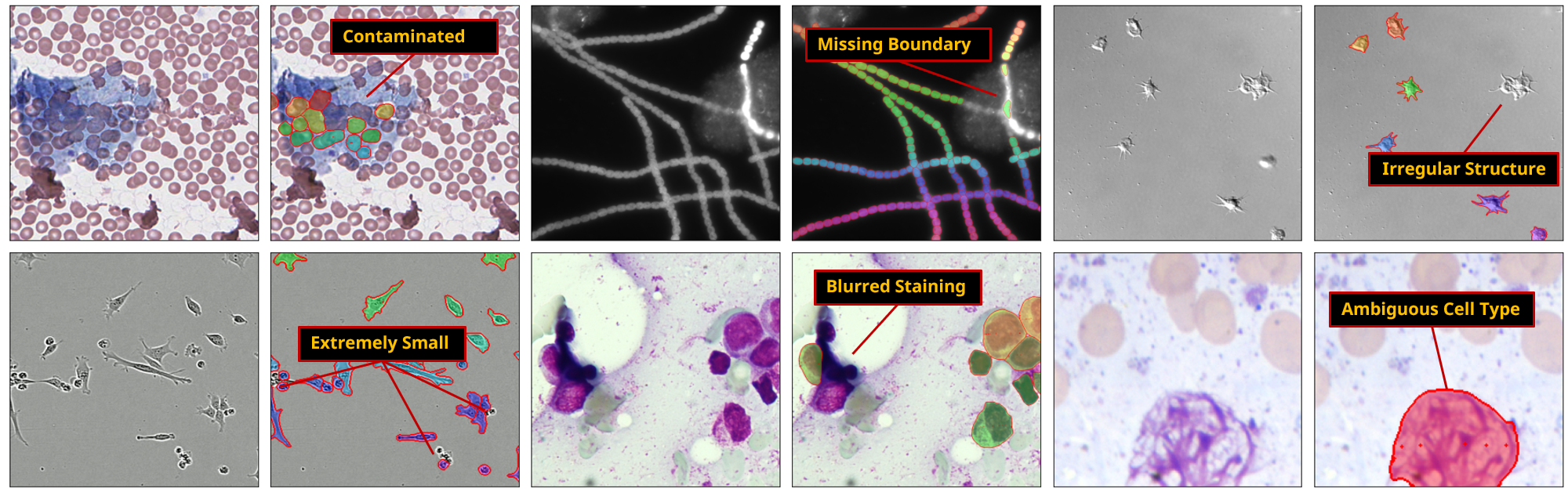}
    \caption{MEDIAR failure cases across different modalities. We zoomed in on the images for clear visualization and categorized the failure types. Note that the magnifications are different.}
    \label{fig:failure_cases}
    \vspace{-5pt}
\end{figure}

\vspace{-2pt}
Although our MEDIAR performs cell instance segmentation surprisingly well in various situations, it suffers from capturing cells in a few cases. We provide the failure cases in \autoref{fig:failure_cases} with the categorized failure types. At first, when the cell regions in microscopy image are distorted, MEDIAR sometimes captures the organisms in the contaminated area as cells (\textit{Contaminated}) or drops the cells in the region (\textit{Missing Boundary, Blurred Staining}). Second, when the cell consists of only a few pixels (\textit{Extremely small}), or the shape has an extraordinarily irregular structure (\textit{Irregular Structure}), the cells are not recognized on occasion. As humans can capture cells even in those cases using their prior knowledge, we expect that integration of prior knowledge may further improve the robustness of prediction. On the other hand, some ambiguous objects are captured as cells (\textit{Ambiguous}). Those cases may result from just a particular type of noise from the staining process or specific cell phenomena (e.g., apoptosis \citep{apoptosis}), which depends on the cell recognition criteria.
\section{Conclusion}
\vspace{-3pt}
This study investigates the difficulties in cell segmentation on multi-modality microscopy images and proposes a robust and generalizable algorithm, MEDIAR, to conduct cell instance segmentation in various situations. To overcome the modality heterogeneity, we harmonize data-centric and model-centric approaches. On the data-centric side, we suggest a strategy to balance modalities during training and a comprehensive pretraining strategy with replaying their knowledge in fine-tuning. On the model-centric side, we propose a model structure to recognize cell regions and identify each cell object with a corresponding efficient inference method. Our MEDIAR shows remarkable success in various microscopy images and identifies the cell instances well across different modalities.

\paragraph{Broader Impact} We believe that automated analysis of microscopy images is a crucial first step for many bio-medical applications. Providing the trained model with open-source code release may facilitate the advance of biomedical research. However, despite the remarkable success of MEDIAR in microscopy images, it sometimes needs to improve in recognizing cells depending on the imaging quality or when the cells in the image have inconsistent sizes or shapes. Although tuning the model weights using additional datasets can be the solution, the bio-medical participators should consider this problem before deploying the method. Furthermore, as the MEDIAR framework does not use unlabeled datasets, how to properly incorporate approaches for unlabeled datasets would be a promising extension for MEDIAR.

\section*{Acknowledgement}
\vspace{-4pt}
For participation in the NeurIPS 2022 Cell Segmentation Challenge, the proposed method in this paper has not used any private datasets other than those provided by the organizers and the official external datasets and pretrained models. The proposed solution is fully automatic without any manual intervention. This work was supported by Institute of Information \& communications Technology Planning \& Evaluation (IITP) grant funded by the Korea government(MSIT) (No.2019-0-00075, Artificial Intelligence Graduate SchoolProgram(KAIST).

\clearpage

\bibliography{main}
\bibliographystyle{plain}

\end{CJK}
\end{document}